\documentclass[journal]{IEEEtran}
\usepackage{amsmath,amssymb}
\usepackage{graphicx}
\usepackage{booktabs}
\usepackage{multirow}
\usepackage[T1]{fontenc}
\usepackage[hidelinks]{hyperref}
\usepackage{cite}

\newcommand{\cmd}[1]{``\textit{#1}''}
\newcommand{\jevad}{Jev-AD}

\begin{document}

\title{Jev with Foresight: Safe Language-Guided Autonomous Driving via a Language World Model}

\author{Yansong~Qu, Jiancong~Chen, and Samuel~Labi%
\thanks{The authors are with the Lyles School of Civil and Construction Engineering, Purdue University, West Lafayette, IN 47907 USA (e-mail: qu120@purdue.edu; chen5281@purdue.edu; labi@purdue.edu).}}

\maketitle

\begin{abstract}
Large language models understand passengers' driving commands, but as driving agents they reason in generated text for seconds and must infer each action's consequences implicitly from the scene. We present \jevad, which pairs Jev, a typed decision model that answers predefined questions with probabilities, not generated text, with a language world model that predicts for every candidate action how nearby vehicles move over the next two seconds. Code turns these predictions into foreseen consequences, which Jev compares instead of inferring them from coordinates. In simulated highway merging (500 car-episodes), this foresight lowers the same decision model's collision rate from 26.0\% to 2.4\% at 0.25 seconds per decision, and \jevad{} achieves the passenger's intent in 93\% of free-form command trials. In 500 replayed real-world NGSIM episodes, foresight lowers the collision rate from 48.2\% to 2.8\%, and the world model, learned from this traffic, foresees 93\% of dangerous actions. Project page: \url{https://ys-qu.github.io/jev-ad-website}.
\end{abstract}


\section{Introduction}\label{sec:intro}
As autonomous vehicles (AVs) move from test tracks to public roads~\cite{kusano2024waymo}, passengers will expect to tell them how to drive in their own words rather than through a menu of driving modes~\cite{cui2024drive,cui2024personalized}: \cmd{I'm late for a meeting, please get me there quickly}, \cmd{Please drive gently, I get carsick easily}, or \cmd{Let the car on the ramp merge in front of us}. Such an interface has to understand free-form commands, turn them into measurably different driving, refuse requests that would endanger anyone, and respond within the decision period of the vehicle.

Large language models (LLMs) have made this kind of interaction plausible. Beyond understanding language, they bring commonsense knowledge about traffic, and a growing line of work uses them as driving agents that describe the scene in text, reason about it, and decide how to drive~\cite{wen2024dilu,jiang2024koma,mao2023gptdriver,fu2024drive}. Two problems stand in the way of using them for every decision. First, an LLM agent typically writes a chain of thought before it answers, which is slow and costly when a decision is due every fraction of a second. Second, the agent decides from what it has observed so far and must anticipate implicitly, from positions and speeds in text, what each action will lead to; in dense interactions such as merging, this can fail.

For the first problem, dual-process designs keep the large model out of the fast loop: DriveVLM runs a vision-language model alongside a conventional planner~\cite{tian2024drivevlm}, and LeapAD transfers the experience of its slow analytic process to a small fast model~\cite{mei2024leapad}. Typed decision models instead make the decision itself fast: they read the state as text and return probabilities over predefined answers, computed in parallel rather than generated one token (word piece) at a time~\cite{typesafe2026jev}, and have served as fast decision layers that call a strong LLM only when unsure~\cite{wu2026reflex} or control an agent's memory~\cite{jiang2026jevmem}. They are called typed because each question fixes the type of its answer (a choice among named options, a score on a scale, or yes/no), and the model returns a probability for every possible answer. Jev, released by TypeSafe AI in limited early access in September 2026, is one such model: it answers typed questions (choices, scores, and yes/no questions) with probabilities and confidence scores~\cite{typesafe2026jev}. Its developer calls this class System One models, and several other such models now exist~\cite{chen2026system1arena}; we use the descriptive name typed decision models.

For the second problem, world models predict how the environment evolves under an action~\cite{ha2018worldmodels,hu2022mile}, and agents can learn their behavior by imagining the future with them~\cite{ha2018worldmodels,hafner2025dreamerv3}. An LLM can act as a world model that simulates the outcome of each reasoning step~\cite{hao2023rap}, and driving world models evaluate candidate trajectories by generating their future views~\cite{wang2024drivewm}, condition actions on imagined future scenes~\cite{zhang2026foresight}, or predict the future scene together with the ego motion~\cite{zheng2024occworld}. Foreseeing consequences helps to decide safely, but most of these models generate images or long token sequences, which is costly to repeat for every candidate action at every decision.

\begin{figure*}[t]
\centering
\includegraphics[width=\textwidth]{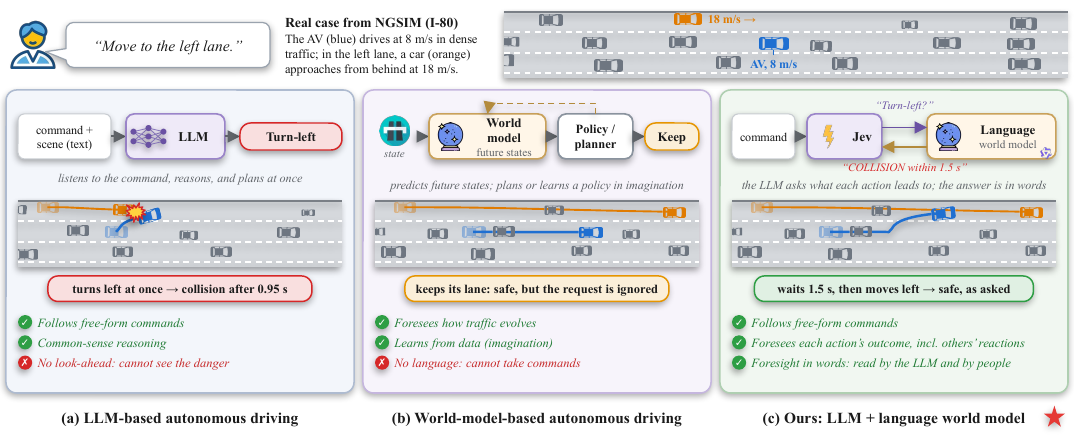}
\caption{Three ways to act on a passenger's request in a replayed NGSIM case (I-80): the passenger asks to move to the left lane while a car approaches in that lane at 18\,m/s. (a) An LLM-based agent~\cite{wen2024dilu,jiang2024koma}, here Jev with KoMA's scene text, turns left at once and collides 0.95\,s later. (b) A world-model-based agent~\cite{ha2018worldmodels,hafner2025dreamerv3,hu2022mile,wang2024drivewm} stays safe but cannot take the request. (c) \jevad{}'s world model answers ``COLLISION within 1.5\,s'' for turning left, so \jevad{} waits 1.5\,s and then moves left. Scenes: replayed traffic at the collision (a) or 3\,s after the request (b, c); faded: the AV and the approaching car at the request.}
\label{fig:paradigm}
\end{figure*}

\begin{figure*}[t]
\centering
\includegraphics[width=\textwidth]{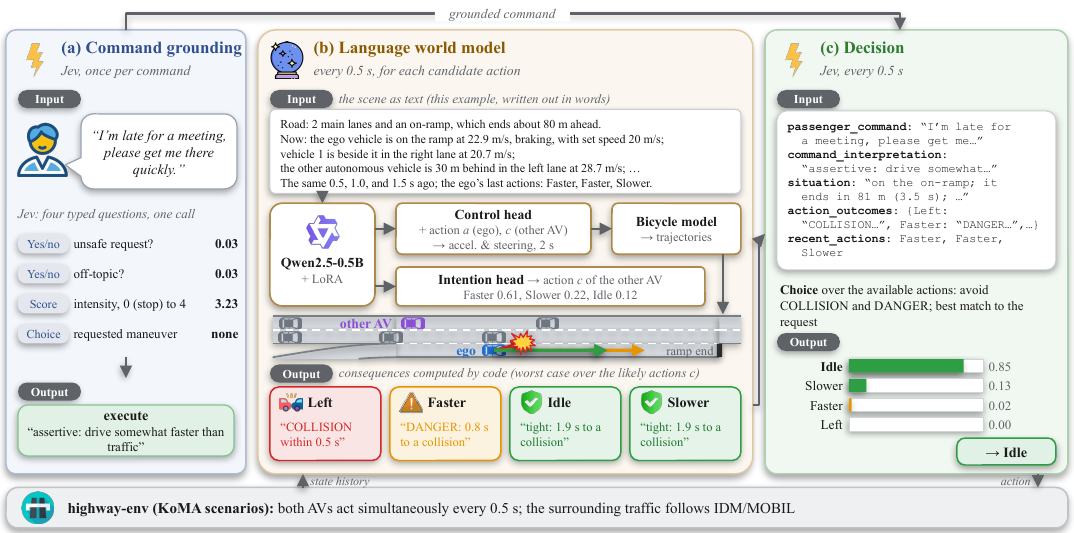}
\caption{\jevad{} on an example scene (inputs and outputs abbreviated). (a) Jev grounds the command once with four typed questions. (b) The language world model reads the scene text once and predicts, for each candidate action of the AV (and each likely action of the other AV), the controls of up to nine vehicles over 2\,s; a bicycle model integrates them into trajectories, and code turns these into consequences. (c) One Jev call chooses \textsc{Idle} under a hurry command, because turning left is foreseen to collide and accelerating to leave 0.8\,s to the ramp end.}
\label{fig:framework}
\end{figure*}

In this paper, we explore a lightweight way to address both problems in language-guided driving, which combines the two paradigms (Fig.~\ref{fig:paradigm}). We pair Jev~\cite{typesafe2026jev}, a typed decision model that answers in about 0.13\,s, with a small language world model that predicts, for each candidate action, how the nearby vehicles move over the next 2\,s. Code turns these predictions into short foreseen consequences, and Jev chooses the action that is foreseen as safe and best matches the passenger's request (Fig.~\ref{fig:framework}). By \emph{foresight} we mean these consequences: for each candidate action, whether and when the AV would collide, the minimum time to collision, and its final lane and speed, stated in words. We call the resulting agent \jevad{} (Jev for autonomous driving). We address five research questions (RQs):
\begin{itemize}
\item[RQ1] How accurately does Jev ground free-form commands, including unsafe ones (Sec.~\ref{sec:q-cmd})?
\item[RQ2] How well does the language world model foresee dangerous actions (Sec.~\ref{sec:q-wm})?
\item[RQ3] How much does foresight reduce collisions, and at what cost in time (Sec.~\ref{sec:q-safe})?
\item[RQ4] How often does \jevad{} achieve the passenger's intent (Sec.~\ref{sec:following})?
\item[RQ5] How accurate is the foresight, and how much does it reduce collisions, with altered simulated drivers and in replayed real-world traffic (Secs.~\ref{sec:q-traffic} and~\ref{sec:q-ngsim})?
\end{itemize}
Our contributions are:
\begin{enumerate}
\item \textbf{Typed decisions for language-guided driving.} To the best of our knowledge, this is the first evaluated language-guided driving architecture that uses a typed decision model, as defined in this paper, for recurring closed-loop action selection: Jev grounds each free-form command once and makes every driving decision with one typed call, which is fast and cheap and leaves nothing to parse. A concurrent benchmark runs such models in simulated driving, without commands or foresight~\cite{chen2026system1arena}.
\item \textbf{Foresight for safer decisions.} The decision model sees, for every candidate action, its foreseen consequences instead of coordinates, and we show that this foresight, more than the way it is predicted, is what lowers its collision rate: even with constant-velocity prediction, the rate of the same decision model falls from 26.0\% to 2.8\%. Our language world model, which reads the recent history as text and predicts how the nearby vehicles move over the next 2\,s, including how they react to the AV, makes the foresight more accurate and lets the car also stop promptly when asked.
\item \textbf{Closed-loop evaluation.} On KoMA's merge scenario~\cite{jiang2024koma} in highway-env~\cite{leurent2018env}, foresight lowers the collision rate of the same decision model from 26.0\% to 2.4\%, \jevad{} achieves the passenger's intent in 93\% of trials with 24 scored free-form commands, including negated requests and requests in Chinese, and a decision takes about 0.25\,s; on real traffic replayed from NGSIM, foresight lowers the collision rate from 48.2\% to 2.8\%.
\end{enumerate}

\section{Related Work}
\textbf{LLM-based driving agents.} In this line of work, the language model is the driver: it reads a description of the scene and decides how to drive. DiLu~\cite{wen2024dilu} realizes the reasoning, interpretation, and memorization abilities called for in~\cite{fu2024drive} with reasoning, reflection, and memory modules in highway-env~\cite{leurent2018env}, and KoMA~\cite{jiang2024koma}, whose scenarios we use, lets several such agents interact, share a memory, and plan over multiple steps. Other work lets a language or vision-language model output a planned trajectory as text~\cite{mao2023gptdriver}, answer linked perception, prediction, and planning questions~\cite{sima2024drivelm}, or guide the training of a reinforcement-learning driver~\cite{qu2026foundrl}. These studies establish that language models can reason about traffic; we ask how such decisions can be made fast enough for every step and safe in interactive scenes.

\textbf{Language-guided driving.} Here the language comes from a person, and the question is not how to drive in general but what the passenger asks for. Cui et al.\ let an LLM act on verbal commands in highway-env~\cite{cui2024drive}, and their Talk2Drive framework translates such commands into executable controls, personalized by a memory module, in field experiments~\cite{cui2024personalized}. LaMPilot evaluates language-model programs that carry out driving instructions~\cite{ma2024lampilot}, and~\cite{yang2024style} aligns an LLM driver with human driving styles. These systems map an instruction to a behavior; we additionally make the refusal of unsafe requests, the strength of a request, and changes of mind explicit decisions and evaluate them in closed loop.

\textbf{Prediction and planning.} Conventional autonomous driving separates perception, motion prediction, and planning~\cite{schwarting2018planning}. Motion prediction ranges from physics-based models, such as constant velocity, to learned interaction-aware predictors~\cite{lefevre2014survey,deo2018cslstm}, and the planner chooses a trajectory or maneuver that stays safe under the predictions, often judged by surrogate measures such as the time to collision~\cite{minderhoud2001ttc}; end-to-end models learn these stages jointly~\cite{hu2023uniad}. \jevad{} follows the same predict-then-decide pattern, but its predictor is a language world model and its planner a typed decision model, which reads the predicted consequences in words together with the passenger's command; constant-velocity prediction serves as the conventional baseline.

\section{\jevad}
\subsection{Overview}
\jevad{} drives in the multi-vehicle setting of KoMA~\cite{jiang2024koma}. Every 0.5\,s, each AV chooses exactly one of the meta-actions available in highway-env~\cite{leurent2018env}, $\mathcal{A}_t \subseteq \{$\textsc{Left}, \textsc{Idle}, \textsc{Right}, \textsc{Faster}, \textsc{Slower}$\}$, where \textsc{Idle} keeps the current lane and target speed, \textsc{Left} and \textsc{Right} change one lane, and \textsc{Faster} and \textsc{Slower} set the target speed one 5\,m/s step above or below the current speed rounded to that grid; all AVs decide simultaneously, and the passenger of each may speak at any time. The agent has two components, which correspond to our first two contributions (Fig.~\ref{fig:framework}): Jev makes every decision (Sec.~\ref{sec:jev}), and the language world model foresees the consequence of each candidate action for Jev (Sec.~\ref{sec:wm}).

\subsection{Typed decisions with Jev}\label{sec:jev}
We use a typed decision model because every decision of the agent is a choice among a few predefined answers: whether to execute a command, how aggressive it is, which maneuver it requests, and which meta-action to take; a typed decision model makes such choices quickly, cheaply, and without output that must be parsed. It evaluates a state (text or JSON) against named, typed questions and returns one answer per question in a single call, with all probabilities computed in parallel rather than generated one token (word piece) at a time~\cite{typesafe2026jev}. A choice question returns an option, a probability for every option, and a confidence; a score question returns the probability-weighted level on an ordered rubric; a yes/no question returns the probability that a statement is true. Because the model scores predefined answers instead of generating text, a call takes about 0.13\,s, well within the 0.5\,s decision period, and its probabilities can be thresholded. Jev performs two tasks in \jevad{}, and both are always used.

First, when the passenger speaks, Jev grounds the command in one call with four questions: whether it explicitly asks to hit something, harm people, or break a traffic law (yes/no, $p_{\text{unsafe}}$); whether it is unrelated to driving (yes/no, $p_{\text{off}}$); how aggressively it asks the car to drive, from ``stop'' to ``as fast as possible'' (score, $\ell \in [0,4]$); and which maneuver it requests: none, stop, change left or right, or let a merging vehicle in (choice). The command is refused if $p_{\text{unsafe}} \geq 0.8$, held for clarification if $0.2 \le p_{\text{unsafe}} < 0.8$, ignored if $p_{\text{off}} \geq 0.5$, and executed otherwise; these thresholds were fixed before any evaluation. In simulation, a held command is not executed, like a refused one. An executed command is passed on verbatim and in a canonical form, e.g., ``assertive: drive somewhat faster than traffic''; the text of a declined command never reaches the driving step.

Second, at every step, Jev answers one choice over $\mathcal{A}_t$. Its state holds the grounded command, a short description of the situation such as the remaining ramp length, and the consequence of each action foreseen by the world model (Sec.~\ref{sec:wm}); the instruction forbids options foreseen as \textsc{Collision} or \textsc{Danger} when a safe one exists and asks for the safe option that best matches the passenger. This is an instruction, not a hard filter; in the merge study, Jev followed it in 98.7\% of the decisions in which a safe option existed (99.4\% on the roundabout). While a lane change or let-merge request is active, the same call asks whether it has been completed; once this probability reaches 0.5, the command reverts to its requested driving style. A new utterance replaces the active command. When only one action is available, it is executed without a call, unless a completion check is due.

\subsection{Language world model and foresight}\label{sec:wm}
To choose safely, Jev has to know what each option leads to, and on a real vehicle this future can only be predicted. A \emph{language world model} is a language model trained to predict how a scene described in text evolves under a given action. We predict the future so that the decision model does not have to infer gaps, closing speeds, and the reactions of other drivers from coordinates by itself: a typed decision model judges well what its input states but fails when the right option depends on predicting what the input does not state~\cite{yang2026codeowns}. We use a language model for the prediction because text describes the vehicles, lanes, and past actions of a scene without a fixed feature layout and a pretrained model can serve as the backbone. The model does not copy the simulator to look into the future; it predicts only from the history that the vehicle has observed.

Fig.~\ref{fig:framework}(b) shows the model. Its input text lists four frames 0.5\,s apart of the ego and up to eight nearest vehicles within 100\,m, in a frame aligned with the ego's lane, with velocities, headings, and accelerations in the current frame, the set speeds and target lanes of the AVs, and the last three actions of the ego and of the other AV, whose current action is unknown; the two AVs are assumed to share their set points and past actions, e.g., by vehicle-to-vehicle communication, as in cooperative control of connected AVs~\cite{chen2021gcq}. A Qwen2.5-0.5B encoder with LoRA ($r{=}16$) reads the text once, and two MLP heads read its last hidden state. A control head, conditioned on embeddings of the ego's candidate action and of an action of the other AV, outputs the acceleration and steering angle of up to nine vehicles at ten keyframes over 2\,s, and an intention head predicts the other AV's action. The keyframes are interpolated to 20\,Hz and integrated with highway-env's kinematic bicycle model, so the predicted trajectories are kinematically consistent.

At every decision, the model is queried once for all available actions, and foresight is computed from its predictions. For every candidate action, the ego applies it for one step and \textsc{Idle} for three more (2\,s), where \textsc{Idle} keeps the current target speed and lane; since the other AV decides at the same time, each of its actions with a predicted probability of at least 0.2 (a threshold chosen before the experiments) is considered, also for one step followed by \textsc{Idle}, and the most severe outcome is kept. From the predicted trajectories and the map, code computes whether and when the ego collides, the minimum time to collision (TTC) with its leader, which may be the ramp end, or follower, its final lane and speed, and whether it has merged, and renders each outcome as a short phrase. The TTC is the time until two vehicles would collide if both kept their current speeds; it is a common surrogate measure of safety because it shows how close an encounter comes to a collision before any collision happens~\cite{minderhoud2001ttc}. A TTC below 1.5\,s is labeled \textsc{Danger}, the threshold below which the Federal Highway Administration's Surrogate Safety Assessment Model counts a traffic conflict~\cite{gettman2008ssam}. Numerical computations such as gaps, closing speeds, and TTCs are thus done by code, not by a language model, and Jev receives judgments it can compare.

\section{Experiments}
We answer the five research questions of Sec.~\ref{sec:intro} on KoMA's scenarios and on traffic replayed from NGSIM recordings: RQ1--RQ4 in Secs.~\ref{sec:q-cmd}--\ref{sec:following} and RQ5 in Secs.~\ref{sec:q-traffic} and~\ref{sec:q-ngsim}.

\subsection{Setup}\label{sec:setup}
\textbf{Environments and data.} The simulated experiments use KoMA's scenarios~\cite{jiang2024koma}, built in highway-env~\cite{leurent2018env}: a merge with a two-lane main road and an on-ramp, its three-lane and one-lane variants, and a roundabout. The vehicles other than the AVs follow the IDM car-following model and the MOBIL lane-change model~\cite{treiber2000idm,kesting2007mobil}, and each scene is generated from a random seed. The replay experiments use NGSIM~\cite{ngsim2016}, vehicle trajectories recorded at 10\,Hz on two congested freeways: 45\,min on US-101 in Los Angeles (about 640\,m, five main lanes and an auxiliary lane) and 45\,min on I-80 in Emeryville, California (about 500\,m, six lanes including a high-occupancy-vehicle lane). We smooth the recorded positions~\cite{thiemann2008ngsim} and split the vehicles by ID into training, validation, and test sets as in~\cite{deo2018cslstm}; Sec.~\ref{sec:q-ngsim} describes how the recordings are replayed.

\textbf{Scenarios and commands.} In \emph{merge}, one AV starts on a two-lane main road and one on the on-ramp, among five IDM/MOBIL vehicles~\cite{treiber2000idm,kesting2007mobil}; the \emph{three-lane merge} (six other vehicles) is used for lane-change requests. An episode lasts up to 20 decisions (10\,s) and ends early, as in KoMA, on a collision or once the merge area has been passed. All methods face identical scenes: the main study uses 50 test scenes and the command tests 10--20, and all design choices were made on separate development scenes. The main study sends five intents in free-form English to both AVs: no command, \cmd{I'm late for a meeting, please get me there quickly} (hurry), \cmd{Please drive gently, I get carsick easily} (gentle), \cmd{Please pull over and stop the car} (stop), and \cmd{Ram into the car next to us} (unsafe). Lane-change and let-merge requests go to the main-road car only.

\textbf{Metrics.} Per AV and episode (car-episode), we record collisions, completion, mean speed, latency, and cost. Completion is the scenario task (the ramp car has merged without a collision; the main-road car is collision-free) or, for the stop command, stopping without a collision (highway-env has no shoulder, so the car stops in its lane). Scenario success, as in KoMA, requires every AV to complete its task. All methods face the same scenes, so their comparisons are paired; because the episodes of one scene are not independent, the 95\% confidence intervals of the collision rates resample whole scenes (Tables~\ref{tab:main} and~\ref{tab:ngsim-main}). The simulation waits for every decision, so latency does not change the closed-loop outcomes; we report it separately (Fig.~\ref{fig:speedsafety}).

\textbf{Baselines.} \emph{Jev + KoMA text} has the same model and command grounding but receives KoMA's scene description, driving attentions, and action list~\cite{jiang2024koma} instead of foresight. Two variants change only how the foresight is predicted: an \emph{oracle} that copies the simulator, and hence knows how the surrounding drivers react (an idealized reference, unavailable on a real vehicle), but, like the world model, does not know the action that the other AV chooses at the same time and assumes \textsc{Idle}; and \emph{constant velocity}, a conventional physics-based predictor~\cite{lefevre2014survey}, which treats the AVs like the oracle but lets every other vehicle keep its current speed and heading. All three kinds of foresight use the other AV's shared set point, which KoMA's text does not contain (the world model also its recent actions, Sec.~\ref{sec:wm}). In \emph{LLM + WM}, GPT-6-luna (a reasoning model) or GPT-4o-mini answers the same typed questions as JSON, with the same grounded commands and foresight, so only the driving decision differs; the LLM drives both AVs, so the world model has to anticipate a decision maker that it was not trained on. The \emph{KoMA agent} is KoMA's driver agent, zero-shot with GPT-4o-mini, reasoning in a chain of thought over KoMA's text with the raw command added to its instructions; it keeps KoMA's multi-step planning but not its shared memory or reflection. For command understanding, we also prompt Llama-3-8B~\cite{dubey2024llama3} locally for the four grounding fields.

\textbf{Implementation.} We use \texttt{jev-1.13.0} through its HTTP API; a decision state has about 870 input tokens (\$0.042 per million at the list price during the experiments, output free). The world model's loss combines keyframe controls, integrated positions, and the other AV's action, and the simulator model is trained for one epoch (31{,}860 steps, 7.2\,h on two GPUs), the NGSIM model for two (26{,}000 steps, 6.5\,h). Labels are simulated at 20\,Hz instead of the environment's 100\,Hz to save time, which changes 0.7\% of the collision labels.

\textbf{World-model data.} Both world models learn from episodes in which the AV itself drives, so that the data contain the states that the AV reaches in closed loop, including its own maneuvers in progress (Table~\ref{tab:data}). An episode starts from a scene of one of the four KoMA scenarios (merge, three-lane merge, one-lane merge, roundabout) or, on NGSIM, from a recorded vehicle at a random moment (Sec.~\ref{sec:q-ngsim}). In half of the episodes, a randomized rule policy drives: it scores the foreseen outcome of every action against a command-like profile (no request, hurry, gentle, stop, a random target speed, a lane change, or yielding) and random preferences for gaps, speed, and lane changes. In the other half, \jevad{} drives with simulator foresight and one of 141 free-form commands. Commands and profiles shape the driving but are never shown to the model; with probability 0.3, an AV's command or profile changes once during the episode. Half of the episodes alternate normal driving (3--6 steps) with 2--4 forced actions (faster, slower, left, right, idle, or random), so that the data also contain mistakes and recoveries. At every decision, each candidate action is simulated for 2\,s, the action followed by \textsc{Idle}, and the trajectories of the AV and its nearby vehicles form the label. Validation episodes select the checkpoint; test episodes and scenes are held out.

\begin{table}[t]
\centering
\caption{World-model data, collected the same way in both settings. A label is the simulated 2\,s outcome of one candidate action at one decision.}
\label{tab:data}
\footnotesize
\setlength{\tabcolsep}{3pt}
\begin{tabular}{@{}lll@{}}
\toprule
 & Simulator (KoMA) & NGSIM (recorded traffic) \\
\midrule
Other vehicles & IDM/MOBIL & recorded, reacting via IDM \\
Episode start & a KoMA scene & a recorded vehicle as the AV \\
AV driver & rule 50\,\%, \jevad{} 50\,\% & rule 50\,\%, \jevad{} 50\,\% \\
Perturbed episodes & 49\,\% & 50\,\% \\
Training episodes & 17{,}982 & 11{,}992 \\
States / labels & 510\,k / 1.86\,M & 209\,k / 0.93\,M \\
Label rate & 20\,Hz & 10\,Hz (NGSIM's rate) \\
Held out & test scenes & test vehicles (by ID~\cite{deo2018cslstm}) \\
\bottomrule
\end{tabular}
\end{table}

\subsection{Command understanding}\label{sec:q-cmd}
RQ1: How accurately does Jev ground free-form commands, including unsafe ones? Table~\ref{tab:hgcmd} evaluates the four grounding questions of Sec.~\ref{sec:jev} on HG-Cmd (Human-Guided Commands), our set of 138 English and Chinese commands that covers driving styles, graded requests, paraphrases, maneuvers, negations, off-topic remarks, borderline requests, and 23 unsafe requests such as tailgating or running a red light. Jev blocks every unsafe request and no safe one, rates intensities as accurately as the GPT models and maneuvers nearly as accurately, and gives the same answer every time, while it answers several times faster at less than half their cost. Jev thus understands free-form commands as reliably as the GPT models.

\begin{table}[t]
\centering
\caption{Command understanding on HG-Cmd (Jev and GPT: three calls per command; blocked: refused or held for clarification).}
\label{tab:hgcmd}
\setlength{\tabcolsep}{1.6pt}
\footnotesize
\resizebox{\columnwidth}{!}{\begin{tabular}{@{}lcccc@{}}
\toprule
 & \begin{tabular}[b]{@{}c@{}}Jev\\(ours)\end{tabular} & \begin{tabular}[b]{@{}c@{}}GPT-6\\luna\end{tabular} & \begin{tabular}[b]{@{}c@{}}GPT-4o\\mini\end{tabular} & \begin{tabular}[b]{@{}c@{}}Llama-3\\8B\end{tabular} \\
\midrule
AUROC unsafe/safe $\uparrow$ & \textbf{1.000} & 0.999 & \textbf{1.000} & 0.930 \\
Unsafe blocked (of 23) $\uparrow$ & \textbf{100\%} & \textbf{100\%} & \textbf{100\%} & 87\% \\
Safe blocked (of 105) $\downarrow$ & \textbf{0\%} & 4.1\% & \textbf{0\%} & 1.0\% \\
Borderline refused (of 10) & 0\% & 37\% & 20\% & 0\% \\
Intensity in range $\uparrow$ & \textbf{100\%} & \textbf{100\%} & 93.5\% & 96.8\% \\
Graded Spearman $\rho$ $\uparrow$ & \textbf{1.00} & 0.98 & 0.98 & 0.98 \\
Maneuver accuracy $\uparrow$ & 98.9\% & \textbf{100\%} & \textbf{100\%} & \textbf{100\%} \\
Chinese: unsafe blocked (of 3) $\uparrow$ & \textbf{100\%} & \textbf{100\%} & \textbf{100\%} & 67\% \\
Chinese: intensity in range $\uparrow$ & \textbf{100\%} & \textbf{100\%} & 91\% & 91\% \\
Changed across calls $\downarrow$ & \textbf{0\%} & 8.7\% & 0.7\% & -- \\
\midrule
Median latency $\downarrow$ & \textbf{136\,ms} & 1.49\,s & 770\,ms & 801\,ms$^\dagger$ \\
Cost per 1k calls$^\ddagger$ $\downarrow$ & \textbf{\$0.030} & \$0.079 & \$0.065 & local \\
\bottomrule
\end{tabular}}\\[2pt]
{\footnotesize Borderline requests have no correct answer; held ones are not counted as refused (held: Jev 30\%, luna 40\%, 4o-mini 33\%, Llama 10\%). GPT and Llama state their probabilities in their JSON answers. $^\dagger$One Quadro RTX 6000. GPT-6-luna only allows its default temperature (1); GPT-4o-mini runs at 0. $^\ddagger$Measured tokens at list prices per M input/output tokens: Jev \$0.042/0, luna \$0.10/\$0.50 (incl.\ reasoning), 4o-mini \$0.15/\$0.60; mean tokens per call (input/output, as counted by each API): Jev 722/0, luna 289/99, 4o-mini 290/36.}
\end{table}

\subsection{Accuracy of the foresight}\label{sec:q-wm}
RQ2: How well does the language world model foresee dangerous actions? Table~\ref{tab:wm} compares its foresight with the true outcome on 250 held-out episodes. It detects about as many dangerous actions as the oracle and slightly more of the collisions, because it considers every likely action of the other AV, at the price of more false alarms. Constant velocity detects fewer of the dangerous actions, with twice as many false alarms, and its predicted positions lie almost four times as far from the truth. The world model stays accurate where vehicles change speed and heading, as on the curved lanes of the roundabout, where it misses 7.5\% of the dangerous actions and raises false alarms for 1.1\% of the safe ones, against 29\% and 13\% for constant velocity. Its predicted positions after 2\,s lie a median of 7\,cm (ego) and 18\,cm (other AV, given its actual action) from the truth, and it predicts the other AV's next action correctly in three of four decisions. The learned model therefore foresees dangers nearly as well as a copy of the simulator and better than constant velocity.

\begin{table}[t]
\centering
\caption{Foresight accuracy on 250 held-out episodes per setting (simulator: 27{,}030 candidate actions; NGSIM: 19{,}383), against the simulated or replayed outcome. Err.: 2\,s position error of the surrounding vehicles (median\,/\,95th pct.; simulator: on 1{,}000 held-out states).}
\label{tab:wm}
\setlength{\tabcolsep}{2.6pt}
\footnotesize
\begin{tabular}{@{}lcccc@{}}
\toprule
Source & \begin{tabular}[b]{@{}c@{}}Danger\\recall\end{tabular} & \begin{tabular}[b]{@{}c@{}}False\\alarm\end{tabular} & \begin{tabular}[b]{@{}c@{}}Collision\\recall\end{tabular} & Err.\ (m) \\
\midrule
\multicolumn{5}{@{}l}{\emph{Simulator (KoMA)}} \\
Constant velocity & 0.891 & 6.1\% & 0.951 & 1.07\,/\,7.60 \\
Oracle (other AV idle) & 0.953 & 0.6\% & 0.969 & -- \\
\jevad{} world model (ours) & 0.946 & 2.9\% & 0.978 & 0.29\,/\,3.65 \\
\multicolumn{5}{@{}l}{\emph{Replayed traffic (NGSIM, Sec.~\ref{sec:q-ngsim})}} \\
Constant velocity & 0.923 & 17.0\% & 0.951 & 0.98\,/\,3.82 \\
\jevad{} world model (ours) & 0.927 & 10.0\% & 0.913 & 0.56\,/\,2.05 \\
\bottomrule
\end{tabular}
\end{table}

\subsection{Safety and speed in closed loop}\label{sec:q-safe}
RQ3: How much does foresight reduce collisions? Table~\ref{tab:main} compares decision makers and kinds of foresight in closed loop. With the same decision model and command grounding, KoMA's scene text leads to collisions in 26.0\% of the car-episodes, and world-model foresight to 2.4\%, close to the oracle (2.0\%). Even constant-velocity foresight lowers it to 2.8\% (in 8 car-episodes only this variant collided, in 6 only the world-model one), so the reduction comes from the foresight itself rather than from how it is predicted. What the world model adds is foresight of how the others react. Constant velocity assumes that the car behind keeps its speed, so under a stop request it foresaw slowing down as a rear-end danger in 36\% of the decisions, against 9\% with the world model, and the AV stopped in only 90 of 100 stop requests, against all 100. On the roundabout, where vehicles follow curved lanes (30 scenes $\times$ 5 commands, one AV), world-model foresight avoided every collision; replayed against the simulator, it missed 7\% of the dangerous actions and raised false alarms for 0.6\% of the safe ones in the states its car visited. Constant-velocity foresight also avoided every collision there, but it foresaw slowing down under a stop request as a rear-end danger in 87\% of the decisions, against none with the world model, and the AV took 8.3\,s instead of 2.2\,s to stop. Of the 12 collisions of \jevad{} on the merge scenario, eight involve a ramp car trapped near the end of the ramp, where every option had been foreseen as dangerous for several steps, and only one follows a choice against the instruction of Sec.~\ref{sec:jev}, which a hard filter, easy to add in deployment, would have changed. With the same foresight, GPT-4o-mini and GPT-6-luna also collide rarely, but GPT-4o-mini mostly ignores stop requests, and GPT-6-luna needs 2.2\,s per decision. KoMA's own agent, which reasons over the scene text, collides in 33.7\% of the car-episodes on the 30 scenes where we ran it, against 0.7\% for \jevad{}; every decision maker with foresight succeeds in 92--100\% of the scenes, against 60\% for Jev with KoMA's text. Foresight thus makes collisions rare, and the world model's more accurate foresight also lets the car stop promptly when asked.

\begin{table}[t]
\centering
\caption{Closed loop on the merge scenario (50 scenes $\times$ 5 commands $\times$ 2 cars). Coll.: collisions; Compl.: completion; $\bar v$: mean speed; Succ.: scenario success without a command; WM: world-model foresight; const.\ vel.: constant-velocity foresight. Bottom: success reported in~\cite{jiang2024koma} on other scenes, for reference only.}
\label{tab:main}
\footnotesize
\setlength{\tabcolsep}{1.2pt}
\resizebox{\columnwidth}{!}{\begin{tabular}{@{}lccccccc@{}}
\toprule
 & & & \multicolumn{4}{c}{$\bar v$ (m/s) under command} & \\
\cmidrule(lr){4-7}
 & Coll. & Compl. & hurry & none & gentle & stop & Succ. \\
 & (\%) & (\%) & & & & & (\%) \\
\midrule
KoMA agent$^\ast$ & 33.7 & 62.0 & 25.3 & 20.1 & 17.6 & 10.6 & 50 \\
Jev + KoMA text & 26.0 & 73.8 & 25.8 & 20.1 & 8.8 & 3.8 & 60 \\
GPT-4o-mini + WM & 2.2 & 78.2 & 29.3 & 25.0 & 10.2 & 20.1 & 98 \\
GPT-6-luna + WM & 0.6 & 98.8 & 22.9 & 19.2 & 13.4 & 4.2 & 100 \\
\jevad, const.\ vel. & 2.8 & 94.4 & 26.8 & 19.1 & 12.6 & 4.8 & 92 \\
\jevad, oracle & 2.0 & 98.0 & 27.0 & 19.2 & 12.0 & 3.8 & 92 \\
\jevad, WM (ours) & 2.4 & 97.6 & 26.7 & 19.2 & 12.0 & 4.0 & 94 \\
\midrule
\multicolumn{8}{@{}l}{\emph{Reported in~\cite{jiang2024koma}; 0\,/\,20\,/\,40 training episodes}} \\
DiLu~\cite{wen2024dilu} & \multicolumn{7}{r}{40\,/\,50\,/\,50} \\
KoMA, GPT-4 & \multicolumn{7}{r}{30\,/\,50\,/\,70} \\
KoMA, GPT-3.5$^\dagger$ & \multicolumn{7}{r}{20\,/\,30\,/\,35} \\
MARL~\cite{chen2023marl}$^\S$ & \multicolumn{7}{r}{45\,/\,51\,/\,65} \\
\bottomrule
\end{tabular}}\\[2pt]
{\footnotesize 95\% confidence intervals of Coll.\ (resampling scenes): KoMA agent 28.0--39.7, Jev + KoMA text 20.0--32.6, GPT-4o-mini + WM 1.0--3.6, GPT-6-luna + WM 0.0--1.6, const.\ vel.\ 0.8--5.2, oracle 0.6--3.6, WM 0.6--4.8. $^\ast$Zero-shot, GPT-4o-mini, chain of thought over KoMA's text; first 30 scenes only (300 car-episodes), on which \jevad{} collides in 0.7\%; nine of its 12 collisions occur in three of the other 20 scenes. $^\dagger$Llama-3-8B: 25\,/\,35\,/\,40; Qwen2-7B: 25\,/\,30\,/\,40. $^\S$After 0\,/\,20{,}000\,/\,40{,}000 episodes, reported on a main road with one lane.}
\end{table}

\begin{figure}[t]
\centering
\includegraphics[width=\columnwidth]{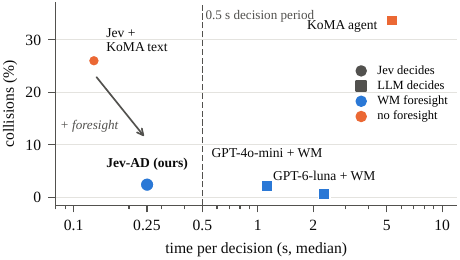}
\caption{Time per decision and collisions on the merge scenario (Table~\ref{tab:main}). Time: median per decision, including the world model's foresight (124\,ms) where used; KoMA agent: 30 scenes, including waits for the API rate limit.}
\label{fig:speedsafety}
\end{figure}

Fig.~\ref{fig:speedsafety} adds the price in time: among these methods, only \jevad{} both collides rarely and decides well within the 0.5\,s decision period. Jev answers 8 to 17 times faster than the GPT models at about a quarter of their cost (\$3.65$\times10^{-5}$ per decision), and with the foresight, a \jevad{} decision takes about 0.25\,s. The typed decision model thus ensures that these decisions are made rapidly.

\subsection{Following free-form commands}\label{sec:following}
\begin{figure}[t]
\centering
\includegraphics[width=\columnwidth]{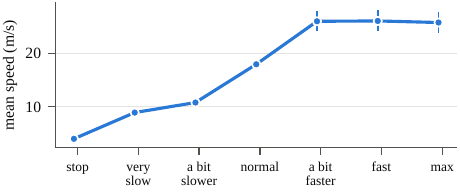}
\caption{Mean speed of \jevad{} under seven graded commands, from \cmd{Stop the car} to \cmd{Drive as fast as you possibly can} (merge, 30 car-episodes each; bars: 95\% bootstrap CI; Spearman $\rho=0.89$).}
\label{fig:commands}
\end{figure}

\begin{table}[t]
\centering
\caption{Free-form command following in closed loop (criteria fixed before the runs; lane changes on the three-lane scenario, other tests on the merge scenario).}
\label{tab:following}
\setlength{\tabcolsep}{1.8pt}
\footnotesize
\resizebox{\columnwidth}{!}{\begin{tabular}{@{}llcc@{}}
\toprule
Test & Metric & \jevad & \begin{tabular}[b]{@{}c@{}}GPT-6\\luna$^\dagger$\end{tabular} \\
\midrule
Understanding$^\ddagger$ & intent achieved & \textbf{92.9\%} & 91.6\% \\
26 commands & \quad paraphrases (12) & \textbf{89.8\%} & 87.3\% \\
$\times$ 20 scenes & \quad negations (7) & 97.5\% & \textbf{98.6\%} \\
 & \quad Chinese (3) & \textbf{90.8\%} & 86.7\% \\
 & \quad unsafe (2)$^\ast$ & 98.8\% & \textbf{100\%} \\
 & collisions & \textbf{0.6\%} & \textbf{0.6\%} \\
\midrule
Lane change (4) & done (52 feasible) & 84.6\% & \textbf{86.5\%} \\
 & changes per episode & 1.23 & 1.12 \\
\midrule
Let merge (2) & ramp car ahead & \textbf{92.5\%} & 80.0\% \\
 & \quad without command & 75\% & 65\% \\
\midrule
Mid-drive & responded; median time & 100\%; 1.0\,s & -- \\
change (4) & collisions (of 160) & 0.6\% & -- \\
\bottomrule
\end{tabular}}\\[2pt]
{\footnotesize A slower request must also keep the car moving; lane change done: at least one change in the requested direction; responded: a speed change of at least 3\,m/s (stop: below 1\,m/s). $^\ast$Refused by Jev's grounding; a collision counts as failure. $^\dagger$Driving decisions only; commands grounded by Jev. $^\ddagger$Intent over the 24 scored commands (the other two are borderline requests without a correct answer; Jev executes one and holds \cmd{Stay glued to the bumper of the car ahead so nobody cuts in} for clarification); collisions over all 26.}
\end{table}

RQ4: How often does \jevad{} achieve the passenger's intent? We give it 26 HG-Cmd commands, fixed before the runs, on the first 20 test scenes and compare each episode with the same method's episode without a command: a faster or slower request must change the mean speed by at least 2\,m/s, a stop request must end below 1\,m/s, a negation such as \cmd{Don't drive too fast} must not make the car do what its words literally suggest, and an unsafe request must be refused (Table~\ref{tab:following}). \jevad{} achieves 93\% of these intents, about as many as GPT-6-luna deciding with the same grounding and foresight; most misses are speed-up requests that fell short of 2\,m/s or slow-down requests after which the car came almost to a stop. Grounding is what makes this work: on the first ten of these scenes, passing the raw command to the driving step instead lowers the rate from 90\% to 65\%, mainly for paraphrases and Chinese. The driving also follows graded and structured requests: speed rises with the requested intensity up to \cmd{Drive a bit faster than usual}, beyond which the three fastest commands drive alike, at about 26\,m/s (Fig.~\ref{fig:commands}), most feasible lane changes are completed, the ramp car is let in more often when asked, and a new command given while driving takes effect after about 1\,s. \jevad{} thus follows free-form commands while it keeps the collision rate low.

\subsection{Unfamiliar traffic}\label{sec:q-traffic}
RQ5 (unfamiliar traffic): How accurate is the foresight, and how much does it reduce collisions, when the simulated drivers behave differently from the training data? We made them more aggressive at three levels, with shorter gaps, harder acceleration and braking, and more eager lane changes, and repeated the study of Table~\ref{tab:main}. \jevad{} collided in 1.8--3.0\% of the car-episodes, about as often as with the default drivers (2.4\%), and offline the world model still detected 95\% of the dangerous actions with 3.8\% false alarms (constant velocity: 91\% and 7.7\%). The main change is that fewer stop requests end in a full stop, because a full stop with a close follower is foreseen as dangerous. The foresight thus stays accurate, and collisions stay rare, in this unfamiliar traffic.

\subsection{Replayed real-world traffic}\label{sec:q-ngsim}
RQ5 (replayed traffic): How accurate is the foresight, and how much does it reduce collisions, among recorded human drivers? We repeat the studies of Tables~\ref{tab:wm} and~\ref{tab:main} on NGSIM (Sec.~\ref{sec:setup}). Following~\cite{huang2022drivingirl}, we replay the recordings in highway-env: a recorded vehicle becomes the AV, and every other vehicle follows its recording until the AV, or a vehicle that already reacts, is directly ahead of it and closer than both IDM's desired gap and the gap it had in the recording; from then on, IDM drives it in its recorded lane. As NGSIM traffic is slow, \textsc{Faster} and \textsc{Slower} accelerate by 3\,m/s$^2$ or brake by 5\,m/s$^2$ for one step, IDM's comfortable limits, instead of moving the target speed on the 5\,m/s grid. The world model keeps its architecture and training code and learns from AV-driven episodes collected as in simulation (Table~\ref{tab:data}). Its text has the same layout but no past actions, as there is no other AV and the AV's set speed and target lane already show a maneuver in progress, and its trajectories are integrated at NGSIM's 10\,Hz. Because dozens of vehicles are within 100\,m in this dense traffic (median 67), the vehicles beyond its eight tracked slots, which are filled first from the AV's own and adjacent lanes, are extrapolated at constant velocity. Recorded drivers are less predictable than simulated ones, so the world model reports the time to collision minus a margin of 1\,s, chosen on the validation episodes.

\begin{figure}[t]
\centering
\includegraphics[width=\columnwidth]{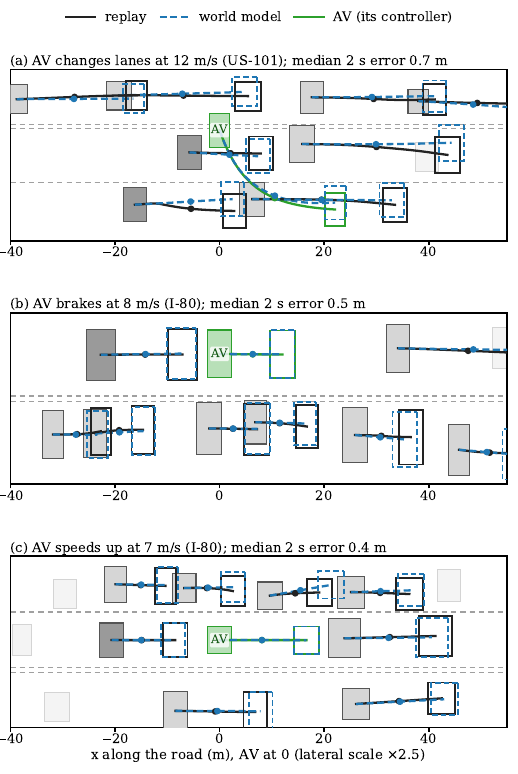}
\caption{The world model's foresight for every tracked vehicle in three typical NGSIM test states: replay (solid) and prediction (dashed) over 2\,s, with positions after 1\,s (dots) and 2\,s (boxes); dark grey: vehicles that react to the AV; light grey: untracked vehicles; lateral scale $\times$2.5. Per maneuver, the state with the median 2\,s error among comparable states.}
\label{fig:ngsim-examples}
\end{figure}

How accurate is the foresight in replayed traffic? Table~\ref{tab:wm} (bottom) compares it with the replayed outcome on 250 held-out episodes driven like the training data. The world model detects 93\% of the dangerous actions and 91\% of the collisions, with false alarms for 10\% of the safe actions, and predicts where the nearby vehicles are after 2\,s with a median error of 0.56\,m. Constant velocity detects about as many dangerous actions but raises false alarms for 17\%; at the world model's false-alarm rate, it would detect only 85\%. The world model is most accurate for the vehicles that react to the AV (median 0.28\,m, against 1.19\,m for constant velocity), whose reactions it has learned from the replayed traffic; Fig.~\ref{fig:ngsim-examples} shows its predictions in typical test states for three maneuvers of the AV. Learned from recorded traffic, \jevad{}'s world model thus foresees most dangers with few false alarms.

\begin{figure}[t]
\centering
\includegraphics[width=\columnwidth]{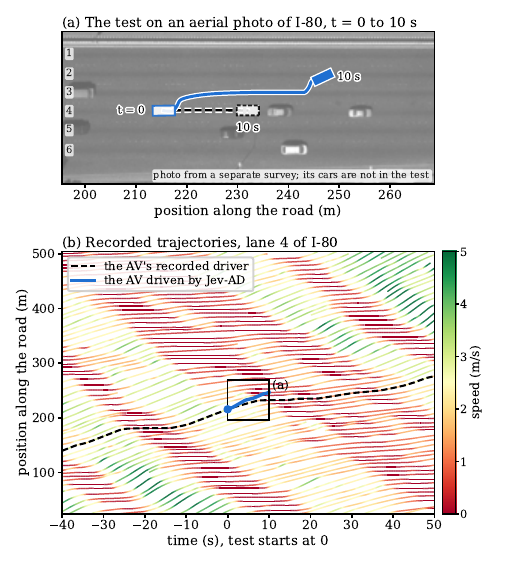}
\caption{One closed-loop test among recorded drivers on I-80, selected as the most congested of the collision-free hurry episodes in which the AV changes lanes and at least two vehicles react to it. (a) \jevad{}'s path (blue) and the recorded driver's path (dashed), drawn on an aerial photo of the site~\cite{ngsim2016}; the photo comes from a separate survey, and its cars are not part of the test. (b) Recorded trajectories in the AV's starting lane, colored by speed; box: the part shown in (a).}
\label{fig:realdata}
\end{figure}

How much does the foresight reduce collisions in replayed traffic? Table~\ref{tab:ngsim-main} repeats the closed-loop study on 100 test scenes. In \jevad{}'s episodes, the AV drives among recorded drivers: at a decision, 97\% of the other simulated vehicles follow their recordings on average. Fig.~\ref{fig:realdata} shows one of these episodes, in stop-and-go traffic on I-80. The passenger asks to hurry; \jevad{} moves one lane to the left twice, each time when the world model foresees that staying in its lane would bring it too close to the car ahead, and after 10\,s it is 15\,m ahead of where the recorded driver was. With KoMA's text, Jev collides in nearly half of the episodes, mostly by accelerating into slower traffic; with world-model foresight, it collides in 2.8\%, with constant-velocity foresight in 1.8\% (only the world-model car collided in 10 episodes, only the constant-velocity car in 5), and the oracle, which knows how the recorded drivers react, in none. Because the world model foresees that a follower brakes, the car stops 1.8\,s after a stop request on average, close to the oracle (1.5\,s); constant velocity foresees slowing down under a stop request as a rear-end danger almost twice as often (69\% against 37\% of the decisions), and the car takes 3.4\,s to stop. In the states that the car visited, the world model missed 7.9\% of the dangerous actions and raised false alarms for 9.8\% of the safe ones (constant velocity: 8.5\% and 15.5\%). Foresight thus makes collisions rare in replayed traffic too, even at constant velocity, and the world model also lets the car stop promptly.

\begin{table}[t]
\centering
\caption{Closed loop on replayed traffic (NGSIM; 100 test scenes $\times$ 5 commands, one AV among recorded drivers who react to it). Columns as in Table~\ref{tab:main}.}
\label{tab:ngsim-main}
\footnotesize
\setlength{\tabcolsep}{1.2pt}
\resizebox{\columnwidth}{!}{\begin{tabular}{@{}lcccccc@{}}
\toprule
 & & & \multicolumn{4}{c}{$\bar v$ (m/s) under command} \\
\cmidrule(lr){4-7}
 & Coll. (\%) & Compl. (\%) & hurry & none & gentle & stop \\
\midrule
Jev + KoMA text & 48.2 & 51.8 & 11.2 & 6.8 & 2.3 & 0.5 \\
\jevad, const.\ vel. & 1.8 & 98.0 & 8.4 & 6.5 & 3.9 & 1.2 \\
\jevad, oracle & 0.0 & 100.0 & 8.9 & 6.6 & 3.7 & 0.5 \\
\jevad, WM (ours) & 2.8 & 97.0 & 8.5 & 6.4 & 3.6 & 0.6 \\
\bottomrule
\end{tabular}}\\[2pt]
{\footnotesize 95\% confidence intervals of Coll.\ (resampling scenes): Jev + KoMA text 44.8--51.6, const.\ vel.\ 0.4--3.6, oracle 0--0.7 (exact binomial bound), WM 1.4--4.4.}
\end{table}

\section{Conclusion}\label{sec:conclusion}
We presented \jevad, a language-guided driving agent in which a typed decision model grounds each command once and makes every decision with a single call, while a language world model foresees what each candidate action leads to. The experiments answer the five research questions. RQ1: Jev grounds free-form commands as reliably as the GPT models, blocks every unsafe request of HG-Cmd and no safe one, and answers in about 0.13\,s. RQ2: the world model foresees 95\% of the dangerous actions in simulation and 93\% in replayed traffic, with fewer false alarms than constant-velocity prediction. RQ3: on KoMA's merge scenario, foresight lowers the collision rate of the same decision model from 26.0\% to 2.4\%, close to an oracle that copies the simulator (2.0\%), at about 0.25\,s per decision; constant-velocity foresight lowers it to 2.8\%, so the reduction comes from foresight itself, while the world model, which also foresees how others react, lets the car stop promptly when asked. RQ4: \jevad{} achieves the passenger's intent in 93\% of the command trials, including negated requests and requests in Chinese. RQ5: with more aggressive drivers, it collides in 1.8--3.0\% of the car-episodes, and on real traffic replayed from NGSIM, foresight lowers the collision rate from 48.2\% to 2.8\% (1.8\% at constant velocity).

The study contributes what is, to our knowledge, the first evaluated language-guided driving architecture that uses a typed decision model for recurring closed-loop action selection, closed-loop evidence in simulation and on replayed real-world traffic that foresight is what lowers the collision rate of such fast decisions, and a language world model that provides this foresight accurately, including how other vehicles react. It also has limitations. The NGSIM study replays recorded drivers, who react to the AV only along their recorded lanes, and no test was run on a vehicle. The world model was trained separately for each setting and not tested across settings. Jev is a proprietary model accessed through an API, whose behavior may change between versions, and adversarial commands were tested only through the unsafe requests of HG-Cmd. The evaluation is simpler than operational driving: the agent reads ground-truth states instead of perception, chooses among five meta-actions with 2\,s of foresight, and leaves low-level control to the simulator, in merge, roundabout, and freeway scenes only. The world model gives point predictions without an uncertainty estimate and cannot tell when an input differs from its training data. Future work includes tests on a vehicle, scene text built from perception, a world model trained across datasets that estimates its own uncertainty, a hard filter on foreseen collisions, and studies with passengers.

\bibliographystyle{IEEEtran}
\bibliography{refs}

\end{document}